\documentclass[10pt,twocolumn,letterpaper]{article}

\usepackage{cvpr}
\usepackage{times}
\usepackage{epsfig}
\usepackage{graphicx}
\usepackage{amsmath}
\usepackage{amssymb}

\usepackage{subfig}
\usepackage{gensymb} 
\usepackage{booktabs}
\usepackage{multirow}
\usepackage{enumitem}
\usepackage{caption}
\usepackage{nccmath}
\usepackage{color}

\usepackage{authblk}

\usepackage{xpatch}
\xpatchcmd{\NCC@ignorepar}{%
\abovedisplayskip\abovedisplayshortskip}
{%
\abovedisplayskip\abovedisplayshortskip%
\belowdisplayskip\belowdisplayshortskip}
{}{}

\setlist{nolistsep}
\setlist[itemize]{leftmargin=*, itemsep=0ex}

\makeatletter
\renewcommand\AB@affilsepx{, \protect\Affilfont}
\makeatother

\usepackage{comment}

\long\def\comment#1{}
\def\xy{\comment}

\usepackage[pagebackref=true,breaklinks=true,letterpaper=true,colorlinks,bookmarks=false]{hyperref}

\cvprfinalcopy 

\def\cvprPaperID{1396} 

\ifcvprfinal\fi
\begin{document}

\title{HPLFlowNet: \textit{H}ierarchical \textit{P}ermutohedral \textit{L}attice FlowNet for\\Scene Flow Estimation on Large-scale Point Clouds}

\author[1,3]{Xiuye Gu}
\author[2]{Yijie Wang}
\author[3]{Chongruo Wu}
\author[3]{Yong Jae Lee}
\author[2]{Panqu Wang}
\affil[1]{Stanford University}
\affil[2]{TuSimple}
\affil[3]{University of California, Davis}


\maketitle

\begin{abstract}
\vspace{-0.1in}
We present a novel deep neural network architecture for end-to-end scene flow estimation that directly operates on large-scale 3D point clouds.
Inspired by Bilateral Convolutional Layers (BCL), we propose novel DownBCL, UpBCL, and CorrBCL operations that restore structural information from unstructured point clouds, and fuse information from two consecutive point clouds.
Operating on discrete and sparse permutohedral lattice points, our architectural design is parsimonious in computational cost.
Our model can efficiently process a pair of point cloud frames at once with a maximum of 86K points per frame.
Our approach achieves state-of-the-art performance on the FlyingThings3D and KITTI Scene Flow 2015 datasets. Moreover, trained on synthetic data, our approach shows great generalization ability on real-world data and on different point densities without fine-tuning.
\end{abstract}


%
%
%
%
%

\vspace{-0.1in}
\section{Introduction} \label{sec:intro}
\vspace{-0.05in}

\begin{figure}[t!]
\centering
\includegraphics[width=1.0\linewidth]{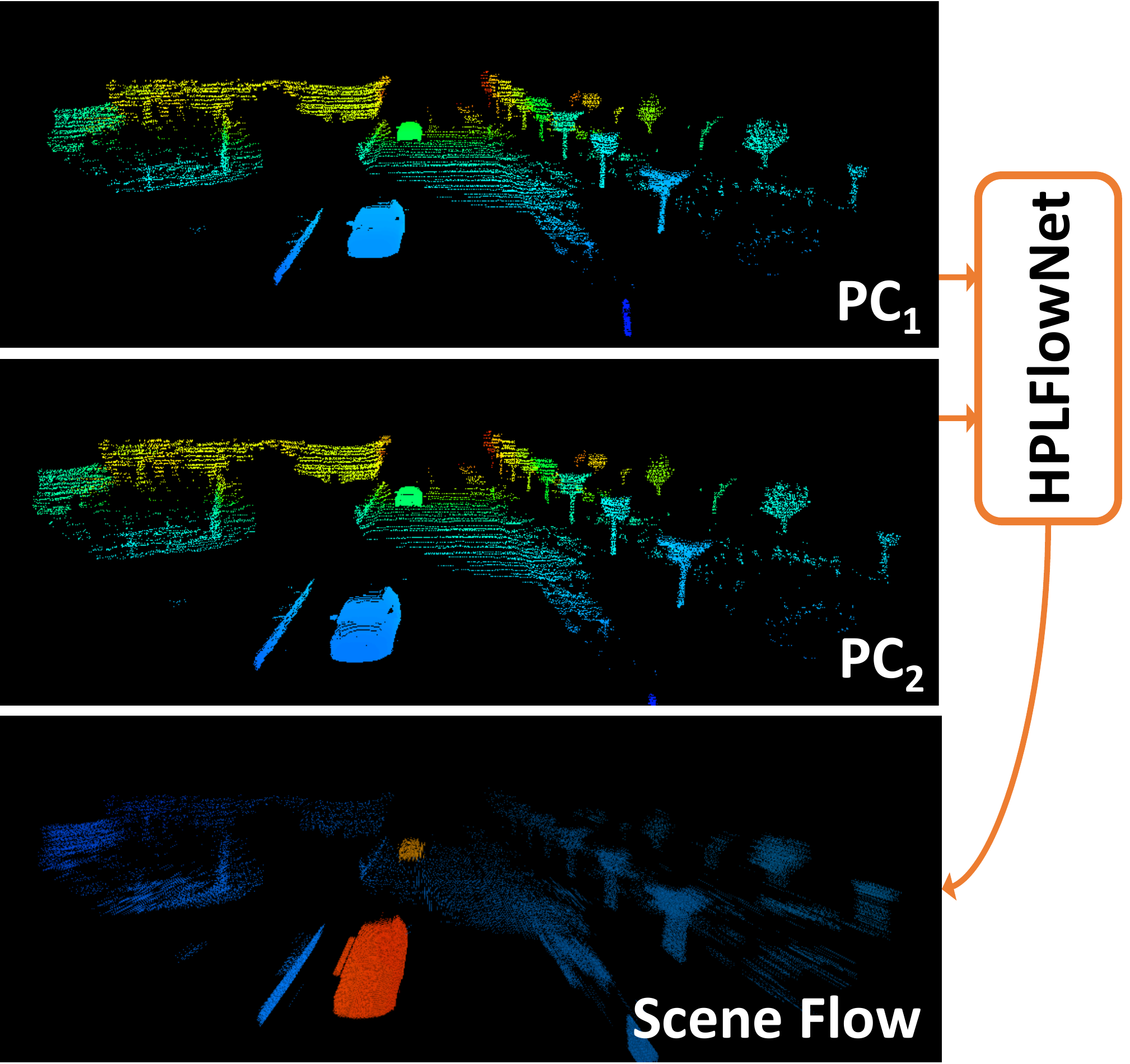}
   \caption{\textbf{Our end-to-end trainable HPLFlowNet} takes two successive frames of point cloud (PC) as input, and outputs dense estimation of the 3D motion field for every point in the first PC frame. The color for scene flow encodes magnitude/velocity from blue to red (small to large).}
\label{fig:intro}
\vspace{-3ex}
\end{figure}

\xy{[1st sentence too similar to FlowNet3D]}
Scene flow is the dense 3D motion field of points. It is the 3D counterpart of optical flow, and is a more fundamental and unambiguous representation -- optical flow is simply the projection of scene flow onto the image plane of a camera \cite{3DSceneFlow}. Scene flow can be useful in various fields, including robotics, autonomous driving, human-computer interaction, and can also be used to complement and improve visual odometry and SLAM algorithms \cite{RGBD_flow, KITTI2015}.

Estimating scene flow in 3D space directly with point cloud inputs is appealing, as approaches that use stereo inputs require 3D motion reconstruction from optical flow and disparities, and thus the optimization is indirect. In this work, we focus on efficient large-scale scene flow estimation directly on 3D point clouds.

The problem statement for scene flow estimation is as follows:  The inputs are two point clouds (PC) at two consecutive frames: $PC_1$ at time $t$ and $PC_2$ at time $t+1$. Generally, each point has an associated feature $f_i = (x_i, y_i, z_i, \ldots) \in \mathbb{R}^{d_f}$, where $(x_i, y_i, z_i)$ are the 3D coordinates for each point. Other low-level features, such as color and normal vectors, can also be included.\footnote{In our experiments, we only use point coordinates to demonstrate the effectiveness of our approach with the bare minimum geometry information.}  The output is the predicted scene flow for each point $i$ in $PC_1$: $\widehat{sf}_i = (dx_i, dy_i, dz_i)$. We use the world coordinate system as the reference system; the goal is to estimate the scene flow of both ego-motion and motion of dynamic objects; see Fig.~\ref{fig:intro}.

Many existing deep learning approaches for 3D point cloud processing \cite{pointnet, pointnet2, kdnet, spectral_5} focus on accuracy but put less emphasis on minimizing computational cost.  Consequently, these networks can only deal with a limited number of points at once due to limited GPU memory, which is unfavorable for large-scale scene analysis.  The reason is twofold: 1) these methods frequently resort to dividing the point cloud into chunks, which can cause global information loss and inaccurate prediction of boundary points due to information loss from the local neighborhood; and 2) these methods also sometimes resort to point subsampling, which impacts performance significantly for regions with sparse point density. \textit{(1) How can we process the entire point cloud of the scene at once while avoiding the above problems}?

Moreover, in \cite{pointnet, pointnet2}, information across multiple points can only be aggregated through max-pooling either globally or hierarchically, and \cite{pointnet2} uses linear search to locate the neighborhood each time. \textit{(2) How can we better restore structural information from unstructured and unordered point clouds}?  Also, in most 3D sensors, the point density is uneven, e.g., nearby objects have larger density while faraway objects have much less density. \textit{(3) How can we make the approach robust under different point densities}?  Finally, scene flow estimation requires combining information from both point clouds.  \textit{(4) How can we best fuse such information}?

\xy{[Maybe make this paragraph more succinct]}
We propose a novel deep network architecture for scene flow estimation that tackles the above four problems. Inspired by Bilateral Convolutional Layers (BCL) \cite{BCN1, BCN2} and the permutohedral lattice \cite{permutohedral_lattice}, we propose three new layer designs: DownBCL, UpBCL, and CorrBCL, which process general unstructured data efficiently (even beyond scene flow estimation).  Our network first interpolates signals from the input points onto a permutohedral lattice. It then performs sparse convolutions on the lattice, and interpolates the filtered signals to coarser lattice points. This process is repeated across several DownBCL layers. In this way, we form a hierarchical downsampling network.  Similarly, our network interpolates the filtered signals from the coarsest lattice points to finer lattice points, and performs sparse convolutions on the finer lattice points. Again, this process is repeated across several UpBCL layers (a hierarchical upsampling network). Finally, the filtered signals from the finest lattice points are interpolated to each point in the first input point cloud. Through the downsampling process, we also fuse signals from both point clouds to the same lattices and perform our correlation operation (CorrBCL). Overall, we form an hourglass-like model that operates on a structured lattice space (except the first and last operation) for unstructured points.

\xy{[Delete using both stereo and point cloud inputs, since they think stereo inputs are not comparable]}
We conduct experiments on two datasets: FlyingThings3D \cite{FlyingThings3D}, which contains synthetic data, and KITTI Scene Flow 2015 \cite{KITTI_2, KITTI_3}, which contains real-world data from LiDAR scans.  Our method outperforms state-of-the-art approaches.  Furthermore, by training on synthetic data only, our model generalizes to real-world data that have different patterns. With a novel normalization scheme for BCLs, our approach also generalizes well under different point densities.  Finally, we show that our network is efficient in terms of computational cost, and it can process a whole pair of KITTI frames at one time with a maximum of 86K points per frame. Code and model are available at \url{https://github.com/laoreja/HPLFlowNet}.

\section{Related work}
\paragraph{3D deep learning.} Multi-view CNNs~\cite{multiview_1, multiview_2, multiview_3, multiview_4, multiview_5} and volumetric networks~\cite{volumetric_1, volumetric_2, volumetric_3, volumetric_and_multiview} leverage standard CNNs with grid-structured inputs, but suffer from discretization error on viewpoint selection and on volumetric representations respectively. PointNet~\cite{pointnet, pointnet2} is the first deep learning approach to work on point clouds directly. Qi~\etal~\cite{pointnet} propose to use a symmetry function for unordered inputs and use max-pooling to globally aggregate information. PointNet++~\cite{pointnet2} is a follow-up with a hierarchical architecture that aggregates information within local neighborhoods. Klokov and Lempitsky~\cite{kdnet} use kd-trees to divide the point clouds and build architectures based on the divisions. Another branch of work~\cite{spectral_1, spectral_2, spectral_3, spectral_4, spectral_5} represent the 3D surface as a graph, and perform convolution on its spectral representation. Su~\etal~\cite{splatnet} propose an architecture for point cloud segmentation based on BCL~\cite{BCN1, BCN2} and achieve joint 2D-3D reasoning.

Our work is inspired by \cite{splatnet}, but with a different focus: \cite{splatnet} focuses on BCL's property of allowing different inputs and outputs to fuse 2D and 3D information in a new way, while we focus on processing large-scale point clouds efficiently without sacrificing accuracy -- which is different from all the above approaches. \xy{[Is the following sentence necessary OR better way to organize?]} In addition, scene flow estimation requires combining information from two point clouds whereas \cite{splatnet} operates on a single point cloud.

\vspace{-10pt}
\paragraph{Scene flow estimation.}
Scene flow estimation with point cloud inputs is underexplored. Dewan \etal \cite{remove_ground_1} formulate an energy minimization problem with assumptions on local geometric constancy and regularization for smooth motion fields. Ushani \etal \cite{remove_ground_2} present a real-time four-step algorithm, which constructs occupancy grids, filters the background, solves an energy minimization problem, and refines with a filtering framework. Unlike \cite{remove_ground_1, remove_ground_2}, our approach is end-to-end.  We also learn directly from data using deep networks and have no explicit assumptions, \eg, we do not assume rigid motions.

Wang \etal \cite{continuous} propose a parametric continuous convolution layer that operates on non-grid structured data and apply this layer to point cloud segmentation and LiDAR motion estimation. However, its novel operator is defined on each point and pooling is the only proposed way for aggregating information. FlowNet3D \cite{flownet3d} builds on PointNet++ \cite{pointnet2} and uses a flow embedding layer to mix two point clouds, so it shares the aforementioned drawbacks of \cite{pointnet2}. Work on scene flow estimation with other input formats (stereo~\cite{flownet3}, RGBD~\cite{VOSF}, light field~\cite{light_field}) is less related, and we refer to Yan~and~Xiang~\cite{sceneflow_survey} for a survey.

\begin{figure*}[tb!]
\centering
\includegraphics[width=1.0\linewidth]{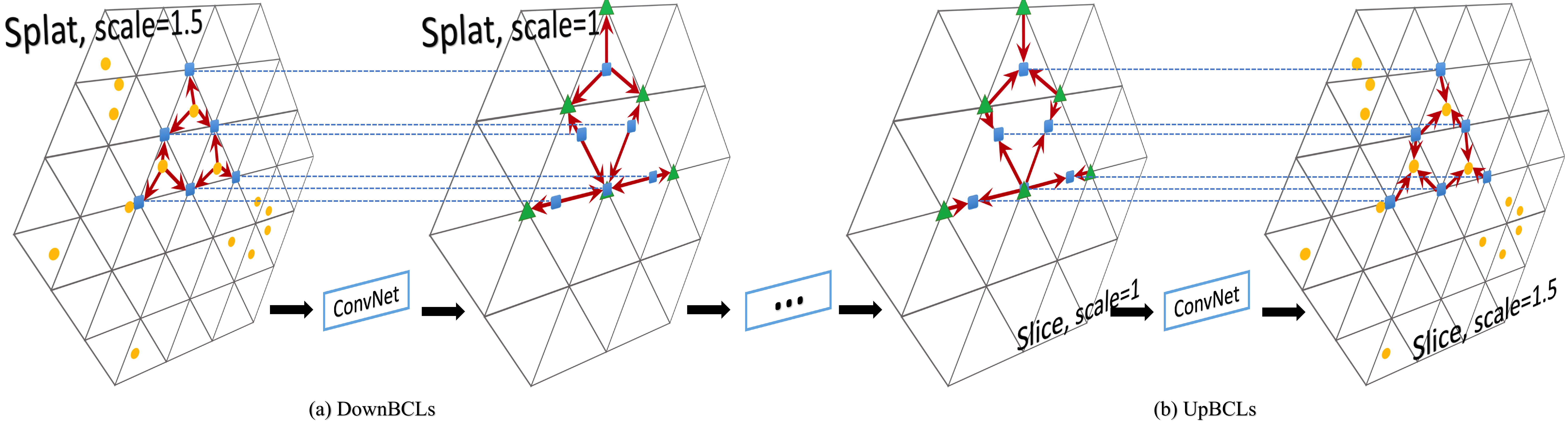}
   \caption{\textbf{Hierarchical DownBCLs and UpBCLs on permutohedral lattice.} DownBCLs are for downsampling and use the Splat-Conv pipeline. During downsampling, the non-empty lattice points (see the blue squares for example) at the previous layer serve as the input points for the next layer on coarser permutohedral lattice, and are splatted onto coarser lattice points (green triangles); vice versa for UpBCLs with Conv-Slice pipeline. \xy{[Change subcaption into regular format]}
   }
\label{fig:BCL}
\vspace{-1ex}
\end{figure*}



\section{BCL on permutohedral lattice}
\label{sec:BCL_review}
\paragraph{Bilateral Convolutional Layer (BCL).}  BCL~\cite{BCN1, BCN2} is the basic building block we use. Similar to how a standard CNN endows the traditional convolution operation with learning ability, BCL extends the fast high-dimensional Gaussian filtering algorithm~\cite{permutohedral_lattice} with learnable weights.

BCL takes general inputs. The convolution is operated on a $d$-dimensional space, and each input point has a
 position vector $p_{in,i} \in \mathbb{R}^d$ and signal value $v_i \in \mathbb{R}^{d_f}$. The position vectors are for locating the points in the defined space on which convolution operates. In our case, $d=3$ and $v_i = p_{in,i}$.

The convolution step of BCL operates on a discrete domain but the input points locate in a continuous domain (for now, without loss of generality, think of the convolution operating on the most commonly used integer lattice $\mathbb{Z}^d$, \ie the regular grid, whose lattice points are $d$-tuples of integers), so BCL: 1) Gathers signals from each input point $p_{in,i} \in \mathbb{R}^d$ onto its enclosing lattice points via interpolation (\textit{splat}), and then 2) Performs sparse convolution on the lattice; since not every lattice point has gathered signals, a hash table is used so that convolution is only performed on non-empty lattice points for efficiency. 3) Returns the filtered signals from each lattice point to the output points inside the lattice point's nearest grids, via interpolation (\textit{slice}); the use of interpolation makes it possible that the output points can locate at different positions from the input points. 
The above procedure forms the three-step pipeline of BCL: \textit{Splat-Conv-Slice}.

\vspace{-10pt}
\paragraph{Permutohedral lattice.} The integer lattice works fine in low-dimensional spaces. However, the number of lattice points each input point interpolates to (\ie, vertices of the Delaunay cell containing each input point)
is $2^d$, which makes the splatting and slicing step have a complexity that is exponential in $d$. Hence, we use the permutohedral lattice\footnote{A lattice is a discrete additive subgroup of a Euclidean space~\cite{permutohedral_properties}. Both regular grid $\mathbb{Z}^d$  and permutohedral lattice $A^*_{d}$ are specific lattices.}  $A^*_d$ \cite{permutohedral_lattice, permutohedral_lattice_thesis, permutohedral_properties} instead: the $d$-dimensional permutohedral lattice is the projection of the scaled regular grid $(d+1)\mathbb{Z}^{d+1}$ along the vector $\vec{1} = [1,...1]$ onto the hyperplane $H_d: \vec{x} \cdot \vec{1} = 0$, which is the subspace of $\mathbb{R}^{d+1}$ in which coordinates sum to zero. The Delaunay cells of the permutohedral lattice are $d$-simplices and the uniform simplices of the lattice tessellates $H_d$. By replacing regular grids with uniform simplices and using barycentric interpolation, the BCL can perform on the permutohedral lattice with the same scheme as on the integer lattice. Special properties of permutohedral lattice make it efficient to compute the vertices of the simplex enclosing any query position and the barycentric weights in $O(d^2)$ time.

 Multiplying the position vectors by a scaling factor $s$, we can adjust the lattice resolution, \ie, larger $s$ corresponds to finer resolution where each simplex contains less points. This effect is the same as scaling the lattice. For better explanation, we interchange the two, and use the term \textit{finer lattice points} and \textit{coarser lattice points}.


\section{Approach: \textit{HPLFlowNet}} \label{sec:arch}

BCL restores structural information from unstructured point clouds, which makes it possible to perform convolutions with kernel size greater than 1. Previous work \cite{splatnet, BCN2} use the same set of input points on the continuous domain for all the BCLs in their network. However, both the time and space cost of splatting and slicing in BCL are linear in the number of input points. \emph{Is there a way to more efficiently stack BCLs to form a deep architecture? How can we combine information from both point clouds for scene flow estimation?} In this section, we address these problems and introduce our HPLFlowNet architecture.

\subsection{DownBCL and UpBCL}

We first introduce the downsampling and upsampling operators, DownBCL and UpBCL. Compared with the three-step operation in the original BCL, DownBCL only has two steps: \textit{Splat-Conv}. The non-empty lattice points at the previous DownBCL become the input points to the next layer, thus saving the slicing step. DownBCL is for downsampling: we stack DownBCLs with gradually decreasing scales, so signals from finer lattice points are splatted to coarser lattice points iteratively, with coarser and coarser resolution and fewer and fewer input points. Similarly, UpBCL, with a two-step pipeline \textit{Conv-Slice}, is used for upsampling with gradually increasing scales.  Signals from coarser lattice points are sliced to finer lattice points directly, thus saving the splatting step. See Fig.~\ref{fig:BCL}.

There are several advantages of DownBCL and UpBCL over the original BCL:

(1) We reduce the three-step pipeline to a two-step pipeline without introducing any new computation, which saves computational cost.

(2) Usually there are much fewer non-empty lattice points than in the input point cloud, especially on the coarser lattice. So we reduce the input size for each DownBCL, except the first one. Similarly, in UpBCL, slicing to the next layer's lattice points instead of to the input point cloud saves computational cost of slicing. In this way, after the first DownBCL and before the last UpBCL, the data size that DownBCLs and UpBCLs have to deal with has nothing to do with the size of the input point cloud, but is instead linear in the number of non-empty lattice points at different scales; i.e., it is only related to the actual volume the point cloud occupies. This is the key advantage of DownBCL and UpBCL that makes computation efficient.

(3) The saved time and memory allow deeper architectures. We use multiple convolution layers with nonlinear activations in between for the convolution step in each DownBCL and UpBCL, instead of the single convolution in the original BCL.

(4) Barycentric interpolation is a heuristic to gather and return signals. The splatting and slicing steps are not symmetric: for input point $i$, let $\mathcal{D}(i)$ denote the vertices of its enclosing simplex; for lattice point $j$, let $\mathcal{V}(j)$ denote the set of input points that lie in a simplex with vertex $j$, $b_{ij}$ denote the barycentric weight used when splatting $i$ to $j$, which is the same weight for slicing $j$ to $i$, and let $g(\cdot)$ denote convolution. Then in the original BCL, the filtered signals for $i$ can be expressed as:
\begin{equation} \label{eq:old_BCL}
v'_i = \sum_{j \in \mathcal{D}(i)} b_{ij}\cdot g(\sum_{k \in \mathcal{V}(j)} b_{kj} \cdot v_k )
\end{equation}	
Even when $g(\cdot)$ is an identity map, we can see that the input signals are changed after the ``identity'' BCL. Also, because of barycentric interpolation, the output signals inside each simplex are always smooth -- this is fine in image filtering~\cite{permutohedral_lattice} where blurring is the expected effect, while it is not ideal for per-point regression, where points within one simplex may have drastically different ground truth. Hence, by removing the slicing step for DownBCL and the splatting step for UpBCL, we reduce such errors caused by the heuristic and asymmetric operations.

\subsection{CorrBCL}
\label{sec:corr}

\begin{figure}[!t]
\centering
\includegraphics[width=1.0\linewidth]{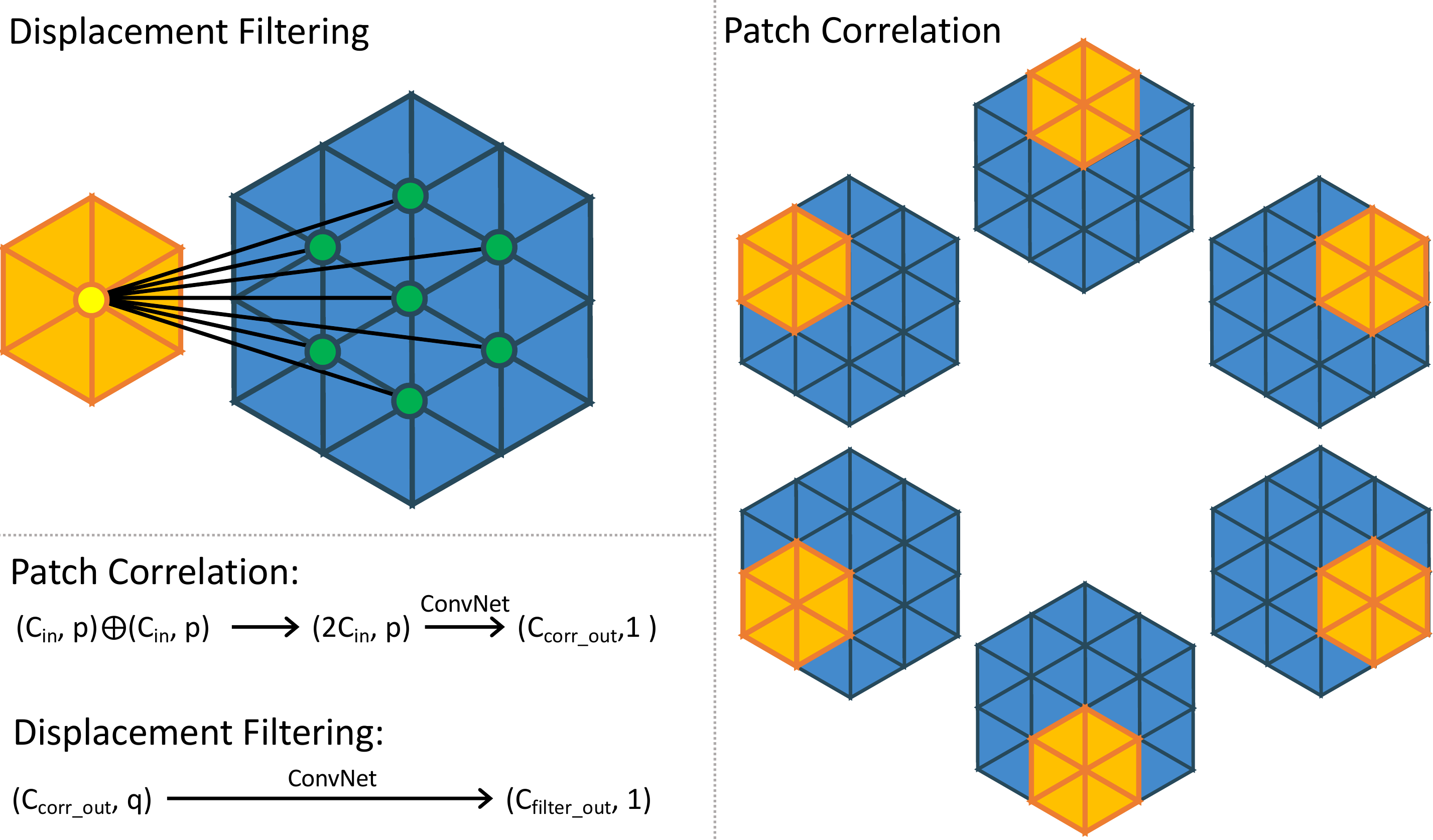}
   \caption{\textbf{Proposed CorrBCL} for combining information from two point clouds, which is crucial for scene flow estimation. The correlation layer consists of two steps: patch correlation and displacement filtering. \xy{[figure adjusted]}
}\label{fig:correlation}
\vspace{-1ex}
\end{figure}

Because of the interpolation design of BCLs, information from two consecutive point clouds can be splatted onto the same permutohedral lattice. In order to fuse information from both point clouds, we propose a novel bilateral convolutional correlation layer (\textit{CorrBCL}), inspired by the matching cost computation and cost aggregation for stereo algorithms \cite{patch}. Our CorrBCL consists of two steps, \textit{patch correlation} and \textit{displacement filtering}.

\vspace{-10pt}
\paragraph{Patch correlation.} Similar to cost matching, patch correlation mixes information from a patch (local neighborhood) at $PC_1$ and another patch at $PC_2$, but in a more general and learnable manner. 

Let $\mathcal{F}_1$ and $\mathcal{F}_2$ denote hash tables storing signals for the two point clouds indexed by lattice positions, $p$ the correlation neighborhood size, and $O_c \in \mathbb{Z}^{p \times d}$ the offset matrix such that  $i^{th}$ neighbor of lattice point at coordinate $x$ is located at $x + O_c[i]$. Then the patch correlation for lattice point in $PC_1$ located at $x$ and lattice point in $PC_2$ located at $y$ is
\begin{equation} \label{eq:patch_corr}
c(x, y) = g\Big(\gamma \big(\mathcal{F}_1(x + O_c[i]), \mathcal{F}_2(y + O_c[i])\big)\mid i=1,...,p\Big)
\end{equation}
where $\gamma(\cdot, \cdot)$ is a bivariate function that combines signals from the two point clouds, and $g$ is a $p$-variate function that aggregates the combined information within each patch neighborhood.

In traditional vision algorithms, $\gamma$ is usually element-wise multiplication, and $g$ is the average function. Our $g$ is instead a convnet, and $\gamma$ is the concatenation function. In this way, we can combine signals of different channel numbers for the two point clouds (element-wise multiplication is unable to do so): we concatenate CorrBCL's output signals and $PC_1$'s signals as input for $PC_1$ and use $PC_2$'s signals only as input for $PC_2$ for the next CorrBCL, see Fig.~\ref{fig:architecture}.

\vspace{-10pt}
\paragraph{Displacement filtering.} Bruteforce aggregation of all possible patch correlation results is computationally prohibitive. Since we are considering point clouds from two consecutive time instances and the $l_2$ norm of the motion is limited, given a lattice point $x$ in $PC_1$, we can move it within a local neighborhood, and match it with the lattice points in $PC_2$ at the moved positions, and then aggregate all such pair matching information for $x$ in a sliding-window manner. This is similar to warping and residual flow in optical flow~\cite{warp_residual_flow_1,warp_residual_flow_2}, but we are warping at every position within the neighborhood.  Let $q$ denote the displacement filtering neighborhood size and $O_f \in \mathbb{Z}^{q \times d}$ denote the offset matrix. For lattice points in $PC_1$ located at $x$, the displacement filtering is defined as:
\begin{equation}
	f(x) = h\big(c(x, x+O_f[j]) \mid j=1,...,q\big)
\end{equation}
where $c(\cdot, \cdot)$ is the patch correlation in Eq.~\ref{eq:patch_corr}, and $h$ is a $q$-variate aggregating convnet.

Note that the whole CorrBCL can be represented as the following general $pq$-variate function:
\begin{equation}
\begin{split}
	\psi(x) = \phi\big(\gamma(\mathcal{F}_1(x + O_c[i]), \mathcal{F}_2(x + O_f[j] + O_c[i]))\\ \mid i=1,...,p, j=1...,q\big)
\end{split}
\end{equation}
We use the factorization technique to save the number of parameters from $O(pq)$ to $O(p+q)$, which is similar to \cite{R2+1D, mobilenets}, and each of our steps has a physical meaning. Fig.~\ref{fig:correlation} shows an example of CorrBCL, where $d=2$ and the correlation and displacement filtering have the same neighborhood size $p=q=7$.

\subsection{Density normalization}

Since point clouds are usually sampled with non-uniform densities and sparse, the lattice points can gather uneven signals.  Thus, a normalization scheme is needed to make BCLs more robust.  All previous work on BCL \cite{BCN1, BCN2, splatnet} use the following normalization scheme following the non-learnable filtering algorithm \cite{permutohedral_lattice}: input signals are filtered in a second round with their values replaced by 1s with a Gaussian kernel, and the filtered values serve as the normalization weights. However, this scheme does not work well for our task (see ablation studies). Unlike image filtering, our filtering weights are learned, and thus it's not suitable to continue using Gaussian filtering for normalization.

We instead propose to add a density normalization term to the splatted signals:
\begin{equation}
u_j = \frac{\sum_{k \in \mathcal{V}(j)} b_{kj} \cdot v_k}{\sum_{k \in \mathcal{V}(j)} b_{kj}}
\end{equation}
where $u_j$ denotes the splatted signals for lattice point $j$, and other notations are the same as Eq.~\ref{eq:old_BCL}.

The advantages of this design are: 1) Normalization is performed during splatting. Compared with the original scheme where the normalization goes through the three-step pipeline, the new scheme saves computational cost. It is worth noticing that \cite{pointnet2} proposes schemes for non-uniform sampling density as well, but their scheme increases computational cost greatly. 2) It applies directly to CorrBCL; and 3) Experiments show that this scheme makes our approach generalize well under different point densities without fine-tuning.

\subsection{Network architecture} \label{subsec:arch}

\begin{figure}[!t]
\centering
\includegraphics[width=1.0\linewidth]{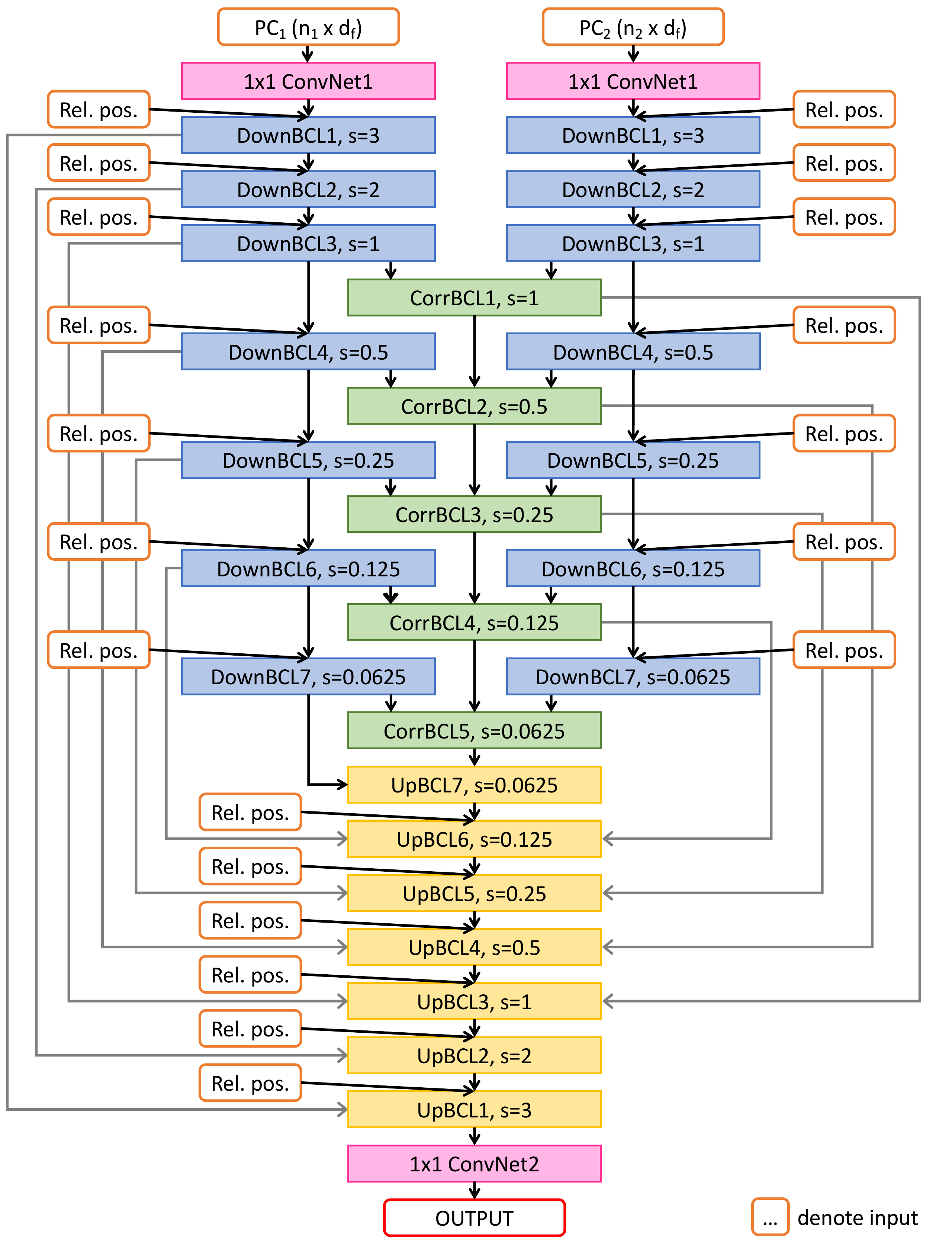}
   \caption{\textbf{HPLFlowNet architecture.} The layers with the same name share weights. $s$ is scaling factor. $Rel.\ pos.$ is explained in Sec. \ref{subsec:arch}.}
\label{fig:architecture}
\vspace{-1ex}
\end{figure}

The network architecture for HPLFlowNet is shown in Fig.~\ref{fig:architecture}. We use an hourglass-like model due to its good performance in applications of 2D images \cite{FCN, unet}. It has a Siamese-like downsampling stage with information fusion and an upsampling stage. In the downsampling stage, DownBCLs with gradually decreasing scales are stacked, so that lattice points in higher layers have larger receptive fields and information within a larger volume is gathered to each lattice point. Since $PC_2$ is important for making scene flow predictions, it goes through all the same layers as $PC_1$ with shared weights. Unlike previous work \cite{flownet3d, flownet} that fuse signals from $PC_1$ and $PC_2$ only once, we use multiple CorrBCLs at different scales for better signal fusion. 
In the upsampling stage, we gradually refine the predictions by stacking UpBCLs of gradually increasing scale, and finally, slicing back to the points in $PC_1$. For each UpBCL, we use skip links from the outputs of their corresponding DownBCL and CorrBCL -- information from different stages can be merged at refining time because layers with the same scaling factor have the same set of non-empty lattice points, 

At each BCL, we concatenate the input signals with its relative positions \wrt its enclosing simplex (its position vector minus the lattice coordinates of its ``first'' enclosing simplex vertex). In Fig.~\ref{fig:architecture}, we use $Rel.\ pos.$ to denote the relative positions. By providing the network with relative positions directly, it can achieve better translational invariance. The CNN we use is translational invariant under certain quantization errors, but unlike standard CNNs, we are interpolating signals from the continuous domain onto the discrete domain, which leads to some positional information loss. By incorporating $Rel.\ pos.$ into the input signals, such loss can be compensated.

Since most layers of our model always operate on sparse lattice points, their computational cost is unrelated to the size of point clouds, but only relates to the actual volume that the point cloud occupies. To train HPLFlowNet, we use the End Point Error (EPE3D) loss: $\|\widehat{sf} - sf\|_2$ averaged over each point, where $\widehat{sf}$ denotes the predicted scene flow vector and $sf$ denotes the ground truth. EPE3D is the counterpart of EPE for 2D optical flow estimation.


\section{Experiments}

We show results for the following experiments: 1) We train and evaluate our model on the synthetic FlyingThings3D dataset, and 2) also test it directly on the real-world KITTI Scene Flow dataset without fine-tuning. 3) We test the model on inputs with different point densities, 4) compare computational cost at both architecture and single-layer level, and 5) conduct ablation studies to analyze the contribution of each component. 

\vspace{-10pt}
\paragraph{Evaluation metrics.}
\textbf{EPE3D} (m): our main metric, $\|\widehat{sf} - sf\|_2$ averaged over each point. \textbf{Acc3D Strict}: a strict version of accuracy, the percentage of points whose EPE3D $<0.05m$ or relative error $<5\%$. \textbf{Acc3D Relax}: a relaxed version of accuracy, the percentage of points whose EPE3D $<0.1m$ or relative error $<10\%$. \textbf{Outliers3D}: the percentage of outliers whose EPE3D $>0.3m$ or relative error $>10\%$. By projecting the point clouds back to the image plane, we obtain 2D optical flow. In this way, we measure how well our approach works for optical flow estimation. \textbf{EPE2D} (px): 2D End Point Error, which is a common metric for optical flow. \textbf{Acc2D}: the percentage of points whose EPE2D $<3px$ or relative error $<5\%$.

\subsection{Results on FlyingThings3D}\label{sec:result_fly}

\begin{table*}
\centering
\footnotesize
\caption{Evaluation results on FlyingThings3D and KITTI Scene Flow 2015. Our method outperforms all baseline methods on all metrics (FlowNet3 is not directly comparable). The good performance on KITTI shows our method's generalization ability.} \label{table:results}
\begin{tabular}{llccccccc}
\toprule
Dataset & Method & EPE3D & Acc3D Strict & Acc3D Relax & Outliers3D & EPE2D & Acc2D \\
\midrule
\multirow{5}{*}{FlyingThings3D} & FlowNet3 \cite{flownet3} & 0.4570 & 0.4179 & 0.6168 & 0.6050 & 5.1348 & \bfseries 0.8125 \\ 
\cline{2-8}
& ICP~\cite{ICP} &0.4062 & 0.1614 & 0.3038 & 0.8796 & 23.2280 & 0.2913\\
& FlowNet3D~\cite{flownet3d} & 0.1136 & 0.4125 & 0.7706 & 0.6016 & 5.9740 & 0.5692 \\
& SPLATFlowNet~\cite{splatnet} & 0.1205 & 0.4197 & 0.7180 & 0.6187 & 6.9759 & 0.5512 \\
& original BCL & 0.1111 & 0.4279 & 0.7551 & 0.6054 & 6.3027 & 0.5669\\
& Ours & \bfseries0.0804 & \bfseries0.6144 & \bfseries0.8555 & \bfseries0.4287 & \bfseries4.6723 & 0.6764 \\
\midrule
\multirow{5}{*}{KITTI } & FlowNet3 \cite{flownet3} &  0.9111 & 0.2039 & 0.3587 & 0.7463 & 5.1023 & \bfseries 0.7803 \\ 
\cline{2-8}
& ICP~\cite{ICP} & 0.5181 & 0.0669 & 0.1667 & 0.8712 & 27.6752 & 0.1056 \\
& FlowNet3D~\cite{flownet3d} & 0.1767 & 0.3738 & 0.6677 & 0.5271 & 7.2141 & 0.5093 \\
& SPLATFlowNet~\cite{splatnet} & 0.1988 & 0.2174 & 0.5391 & 0.6575 & 8.2306 & 0.4189 \\
& original BCL & 0.1729 & 0.2516 & 0.6011 & 0.6215 & 7.3476 & 0.4411\\
& Ours & \bfseries0.1169 & \bfseries0.4783 & \bfseries0.7776 & \bfseries0.4103 & \bfseries4.8055 & 0.5938\\
\bottomrule
\end{tabular}
\vspace{-3.5ex}
\end{table*}

\begin{figure*}[!thb]
\centering
\subfloat{
\includegraphics[width=.24\textwidth]{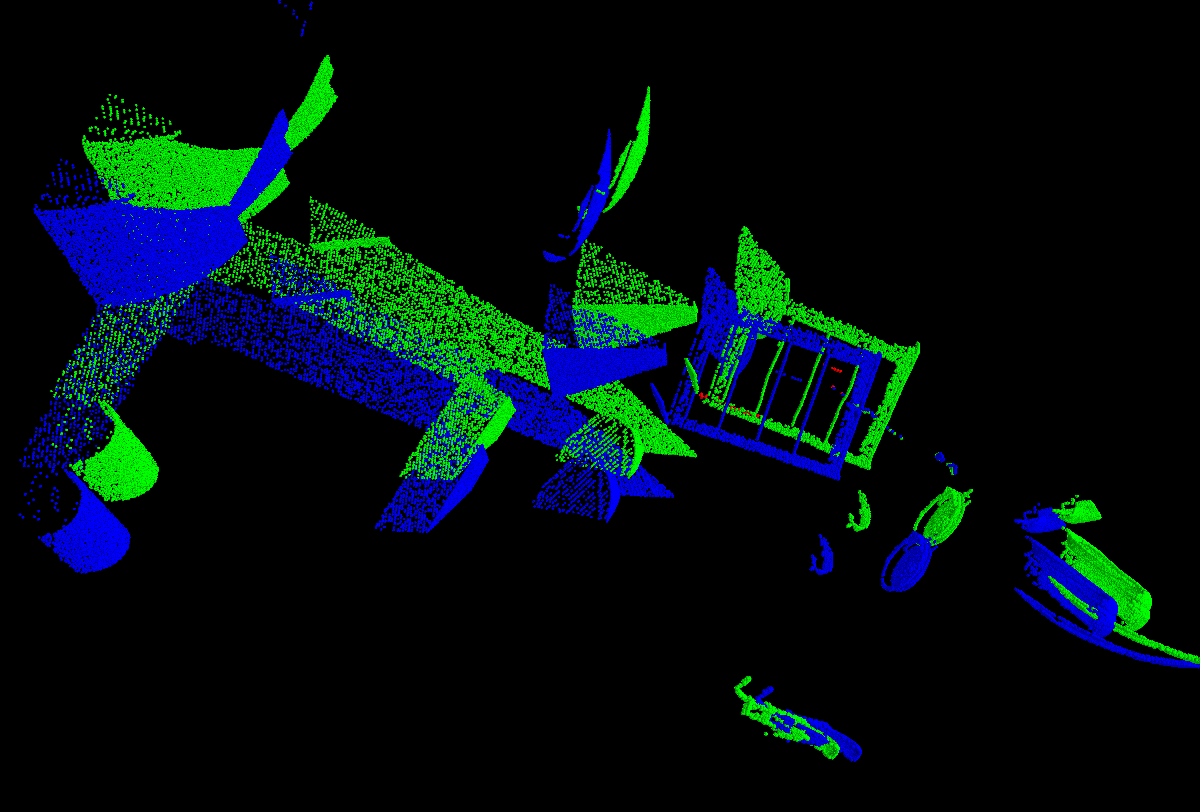}
}\hspace{-0.5em}
\subfloat{
\includegraphics[width=.24\textwidth]{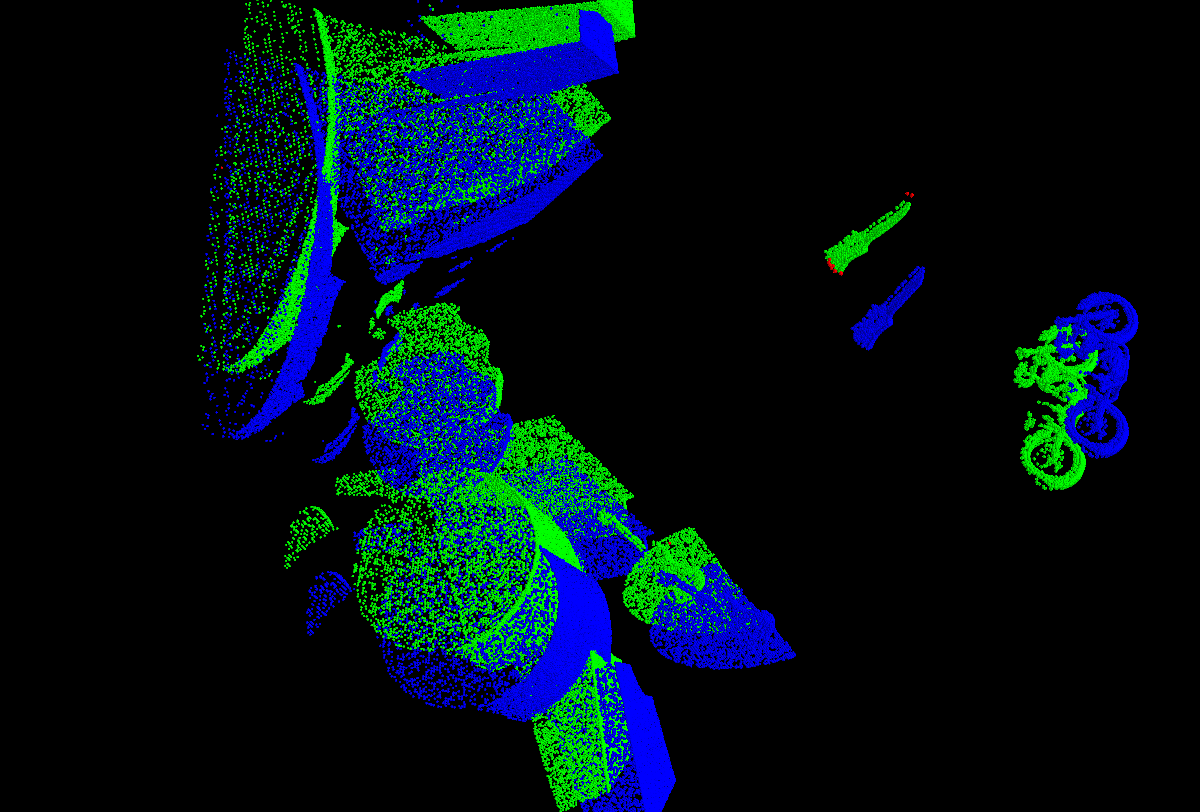}
}\hspace{-0.5em}
  \subfloat{
  \includegraphics[width=.24\textwidth]{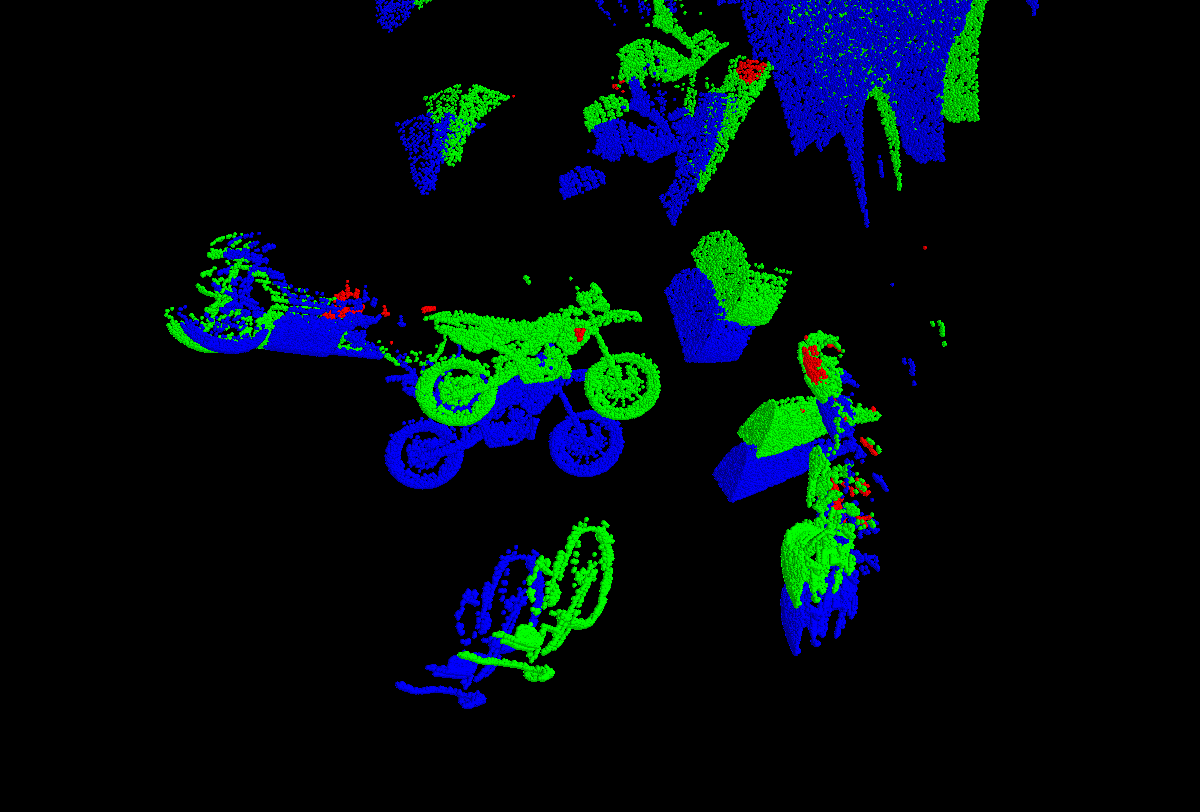}
  }\hspace{-0.5em}
  \subfloat{
  \includegraphics[width=.24\textwidth]{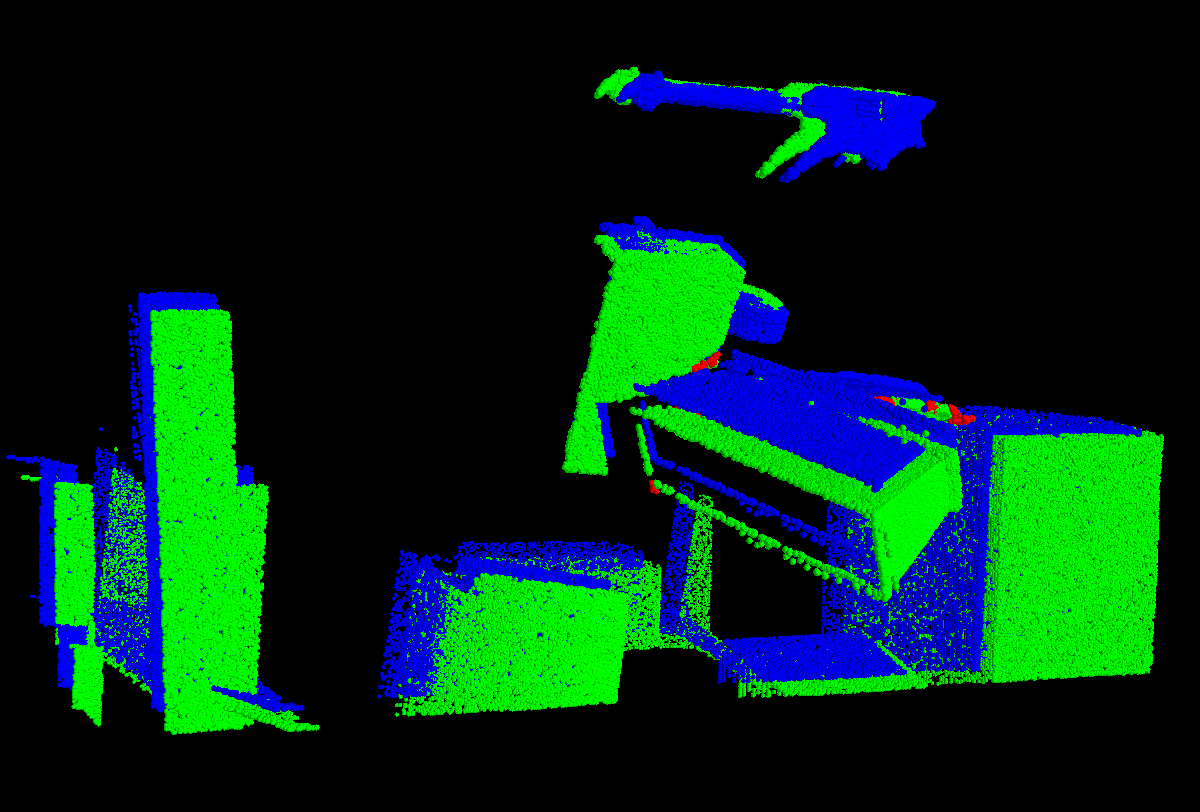}
  }
  \vspace{-1.9ex}

\subfloat{
\includegraphics[width=.24\textwidth]{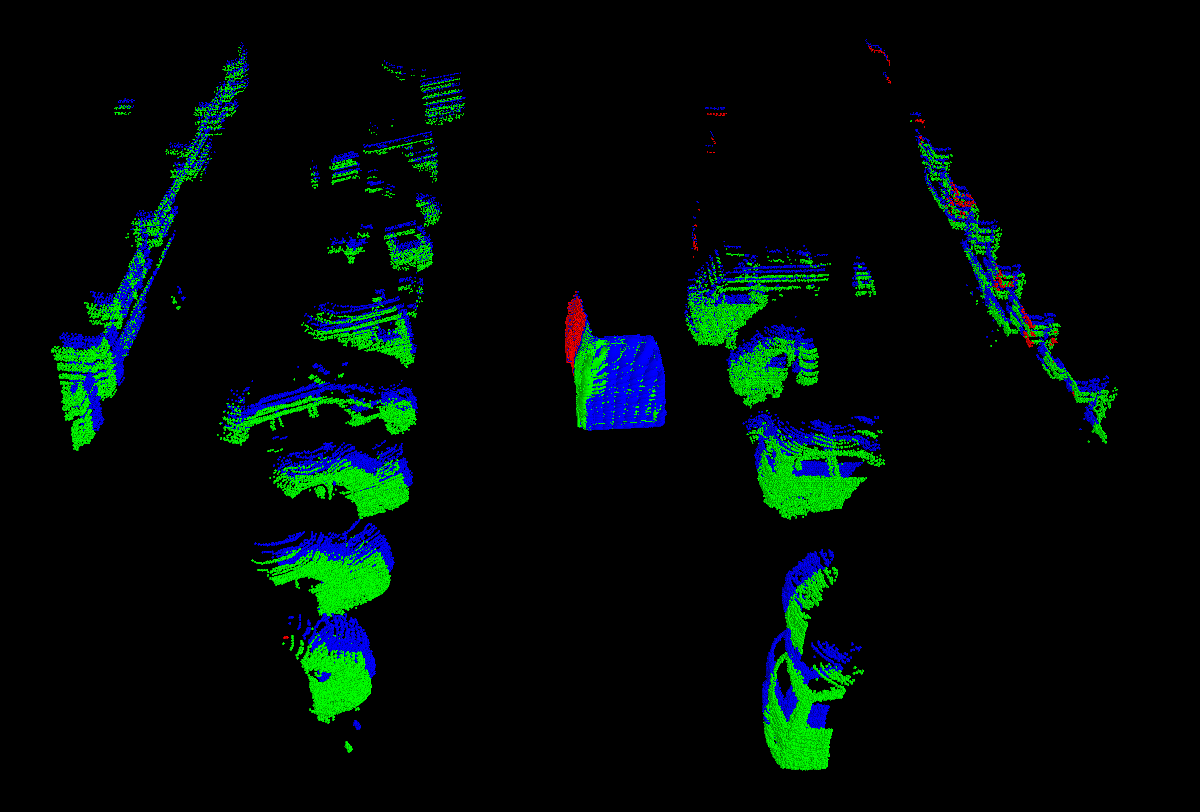}
}\hspace{-0.5em}
\subfloat{
\includegraphics[width=.24\textwidth]{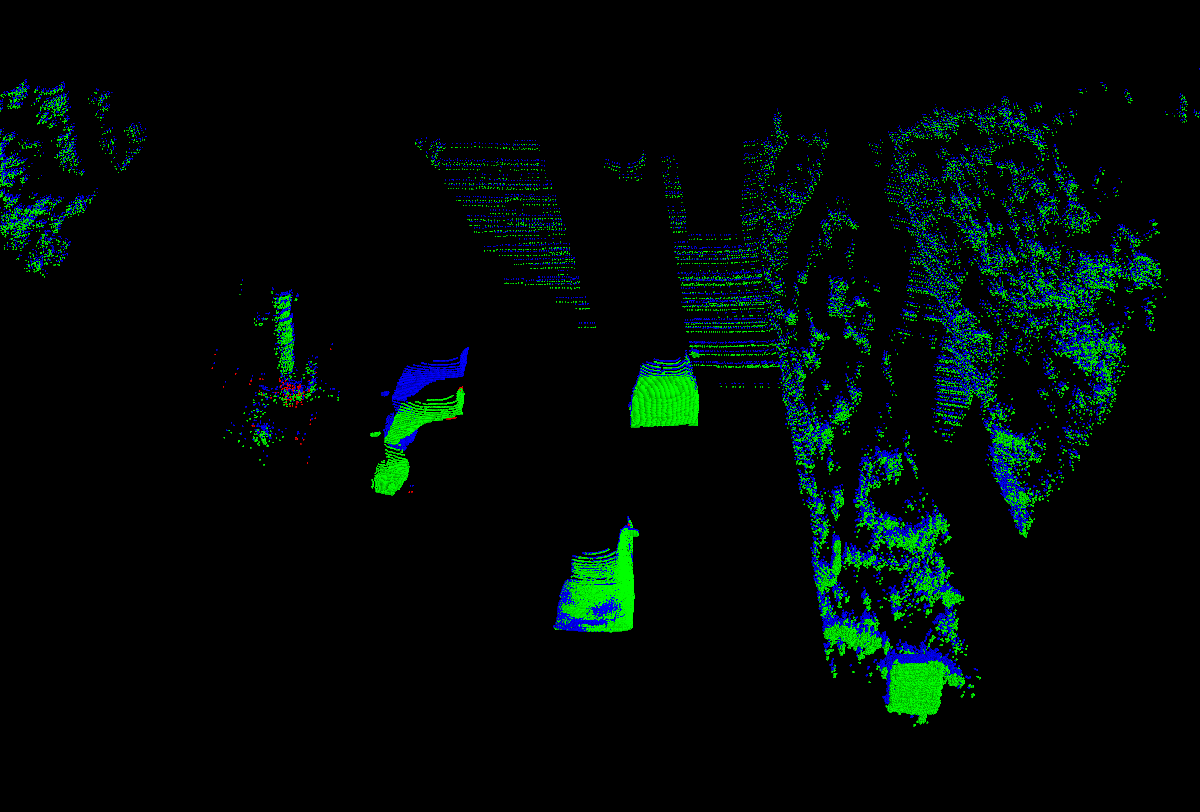}
}\hspace{-0.5em}
\subfloat{
\includegraphics[width=.24\textwidth]{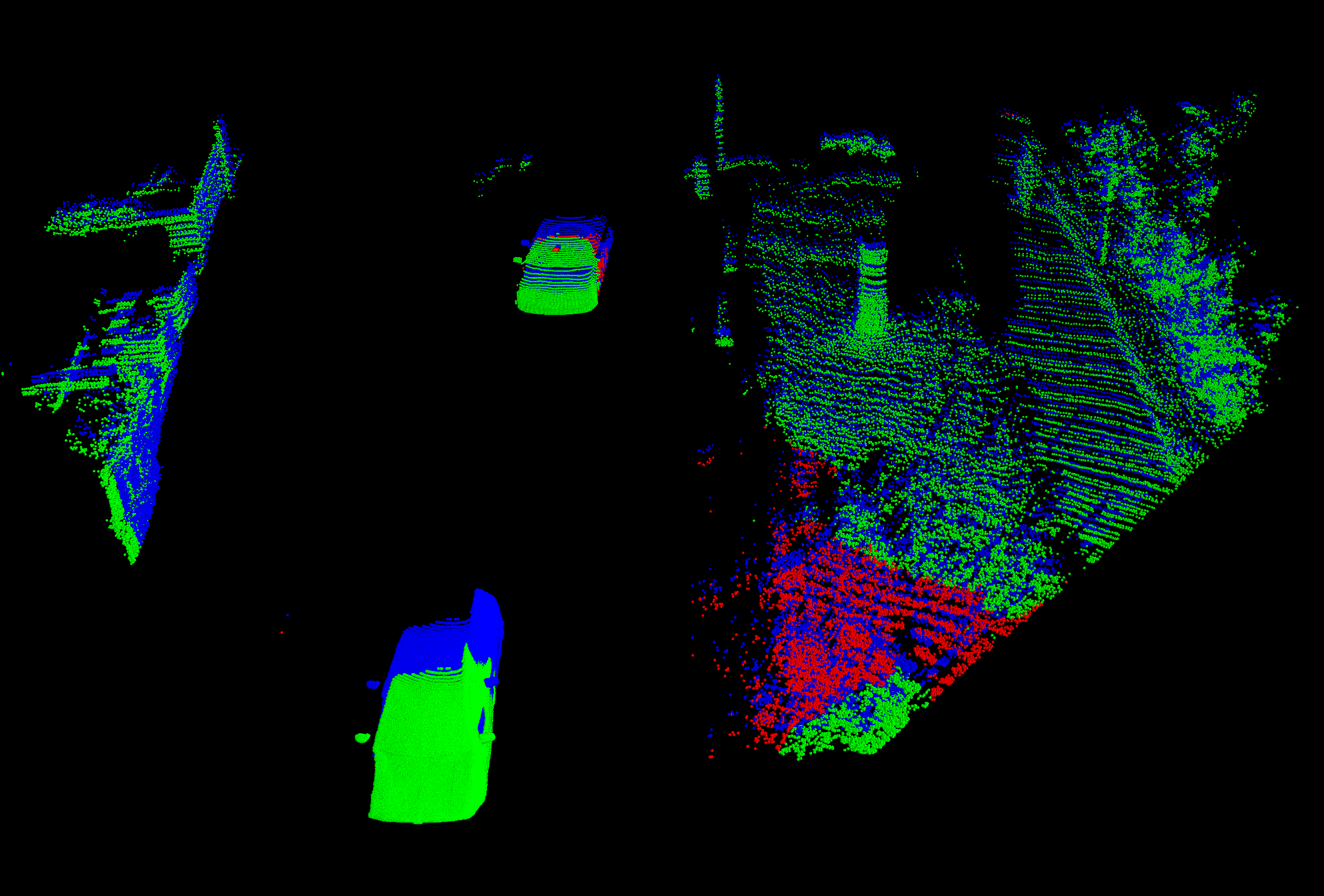}
}\hspace{-0.5em}
\subfloat{
\includegraphics[width=.24\textwidth]{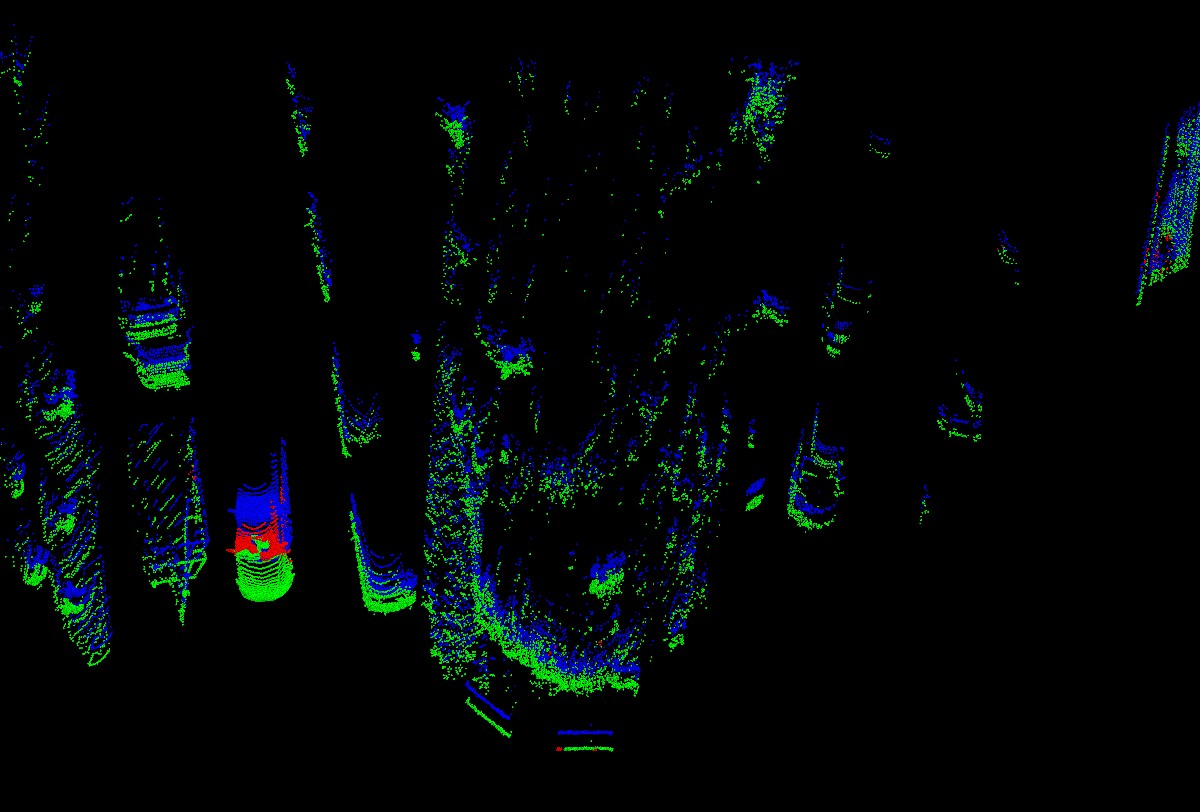}
}
  \caption{\textbf{Qualitative results on FlyingThings3D (top) and KITTI (bottom).} Blue points are $PC_1$, green points are correctly predicted (measured by Acc3D Relax) flowed points $PC_1 + \widehat{sf}$, and red points are ground-truth flowed points $PC_1 + sf$ which are not correctly predicted.
  Note that the objects in the two datasets have very different motion patterns, which shows our method's generalization ability.
  The third figure of the second row shows that some failures are on the ground, which suggests the performance on KITTI may be further improved by better ground removal algorithms.
  }
  \vspace{-1ex}
\label{fig:qualitative}
\end{figure*}

FlyingThings3D \cite{FlyingThings3D} is the first large-scale synthetic dataset that enables training deep neural networks for scene flow estimation.  To our knowledge, it is the only scene flow dataset that has more than 10,000 training samples. We reconstruct the 3D point clouds and ground truth scene flow using the provided camera parameters.

\paragraph{Training and evaluation details.}
Following \cite{FlyingThings3D, flownet2, flownet3}, we use the dataset version where some extremely hard samples are removed\footnote{\url{https://lmb.informatik.uni-freiburg.de/data/FlyingThings3D_subset/FlyingThings3D_subset_all_download_paths.txt}}. To simulate real-world point clouds, we remove points whose disparity and optical flow are occluded.  Following \cite{flownet3d}, we train on points with depth less than 35 meters. Most foreground moving objects are within this depth range. We randomly sample $n$ points from each frame in a non-corresponding manner:  corresponding points for the first frame may not necessarily be found in the sampled points of the second frame. We use $n=8,192$ for training.  To reduce training time, we use one quarter of the  training set (4910 pairs), which already yields good generalization ability. The model finetuned on whole training set achieves 0.0696/0.1113 EPE3D on FlyingThings3D/KITTI. We evaluate on the whole test set (3824 pairs).

\vspace{-10pt}
\paragraph{Baselines.}

We compare to the following methods:

\textbf{Iterative Closest Point}~\cite{ICP}: a common baseline for scene flow estimation, the algorithm iteratively revises the rigid transformation needed to minimize the error metric.

\textbf{FlowNet3D}~\cite{flownet3d}: the state-of-the-art for scene flow estimation with point cloud inputs. Since code is unavailable, we use our own implementation. 

\textbf{SPLATFlowNet}: a strong baseline based on SPLATNet~\cite{splatnet}; architecture is the Siamese network of SPLATNet with CorrBCLs that is about the same depth as our model. It does not use the hourglass architecture, but concatenates all outputs from the BCLs and CorrBCLs of different scales to make the prediction.

\textbf{Original BCL}: We replace DownBCL and UpBCL with the original BCL used in previous work~\cite{BCN1, BCN2, splatnet} while keeping everything else the same as our model.

We also list results of \textbf{FlowNet3}~\cite{flownet3} for reference purposes, since the inputs are in different modalities. It's the state-of-the-art with stereo inputs. We remove points with extremely wrong predictions (\eg, disparity with opposite signs) -- the extremes will induce too much error.

\vspace{-10pt}
\paragraph{Results.}
Quantitative results are shown in Table \ref{table:results}.  Our method outperforms all baselines on all metrics by a large margin, and is the only method with EPE3D below $10cm$. FlowNet3 has the best Acc2D because its optical flow network is optimized on 2D metrics; but it has worse EPE2D since we mainly evaluate on foreground objects, which can have large motions in 2D due to projection and is thus hard to predict. The fact that it is easily affected by extremes (worse EPE3D and EPE2D) also shows that using stereo inputs is more sensitive to prediction errors due to its indirect 3D representation. The reason that our method outperforms FlowNet3D is likely that we better restore structural information and design a better architecture for combining information from both point clouds. Our method and SPLATFlowNet have similar depth and use the same building blocks, so our performance gain can be credited to our hourglass-like model and the skip links that combine filtered signals in the downsampling and upsampling stages. Comparison with the original BCL shows that we improve performance by reduction and verifies that the heuristic and asymmetric nature of the barycentric interpolation makes it better to avoid unnecessary operations. Fig.~\ref{fig:qualitative} shows qualitative results. Our model performs well for complicated shapes, large motions, and also the hard case where multiple neighboring objects have different motions.

\subsection{Generalization results on real-world data}

Next, to study our model's generalization ability to unseen real-world data, we take our model which was trained on FlyingThings3D, and without any fine-tuning evaluate on KITTI Scene Flow 2015 \cite{KITTI_2, KITTI_3}.

\vspace{-10pt}
\paragraph{Evaluation details.}
KITTI Scene Flow 2015 is obtained by annotating dynamic scenes from the KITTI raw data collection using detailed 3D CAD models for all vehicles in motion.  Since disparity is not given for the test set, we evaluate on all 142 scenes in the training set with publicly available raw 3D data, following \cite{flownet3d}. Since in autonomous driving, the motion of the ground is not useful and removing ground is a common step~\cite{remove_ground_1, remove_ground_2, flownet3d}, we remove the ground by height ($<0.3m$). We use similar preprocessing as in Sec.~\ref{sec:result_fly} except that we do not remove occluded points.

\vspace{-10pt}
\paragraph{Results.}
Our method again outperforms all other methods in all metrics by a large margin; see Table \ref{table:results}.  This demonstrates our method's generalization ability to new real-world data. 
Without ground removal, Ours/FlowNet3D EPE3D is 0.2366/0.3331, so ours is still better.  Qualitative results are shown in Fig.~\ref{fig:qualitative}.  Even though our approach is trained on a dataset with very different patterns and different objects, it makes precise estimations in driving scenes where ego-motion is large and multiple dynamic objects have different motions.  It also correctly predicts the trees and bushes which are never seen by the network during training.

\begin{table}
\centering
\footnotesize
\caption{Efficiency comparison: average runtime (ms) on FlyingThings3D measured on a single Titan V. Ours and Ours-shallow are more efficient. 
}\label{table:arch_efficiency}
\begin{tabular}{@{}l|ccc}
\toprule
Method & 8,192 & 16,384 & 32,768\\
\midrule
FlowNet3D~\cite{flownet3d} & 130.8 & 279.2 & 770.0 \\
Ours & 98.4 & 115.5 & 142.8 \\
Ours-shallow & 50.5  & 55.1 & 63.7 \\
\bottomrule
\end{tabular}
\vspace{-1ex}
\end{table}

\subsection{Empirical efficiency}
Our architecture is optimized for performance. To show how efficient our proposed novel BCL variants can be, we make a shallower version \textbf{Ours-shallow} by removing Down/UpBCL6/7 and CorrBCL4/5, and cutting down convolutions (see supp.\ for details).
Table \ref{table:arch_efficiency} shows the efficiency comparison results among different models. Ours is faster than FlowNet3D.
Ours-shallow is very fast and also outperforms all other methods (Table.~\ref{table:point_density_flyingthings3d}). And our runtime does not linearly scale with the number of input points, which empirically validates our architectural design.


We also compare with the original BCL \wrt layer efficiency.
We measure runtime of each BCL variant in our architecture, averaged on FlyingThings3D. We then replace them with original BCLs and do the same. Runtime ratio of ours to original BCL averaged over all layers: $56\%$. We include a more detailed analysis in supp. 


\begin{table}
\centering
\footnotesize
\caption{Results (EPE3D) under different point densities on FlyingThings3D and KITTI. Some results for FlowNet3D are missing since memory runs out without significant sacrifice in speed and/or optimization for memory. Our density normalization scheme works well and achieves superior performance for all testing densities different from the training density.}\label{table:point_density_flyingthings3d}
\begin{tabular}{@{}l@{ }l@{ }c@{\hspace{2ex}}c@{\hspace{2ex}}c@{\hspace{2ex}}c@{}}
\toprule
Dataset & \# points & Ours & No Norm & Ours-shallow & FlowNet3D\\
\midrule
\multirow{4}{*}{FlyingThings3D} 
& 8,192  & 0.0804          & \bfseries0.0790 & 0.0957 & 0.1136 \\
& 16,384 & \bfseries0.0782 & 0.0779 & 0.0932 & 0.1085 \\
& 32,768 & \bfseries0.0774 & 0.0874 & 0.0925 & 0.1327 \\
& 65,536 & \bfseries0.0772 & 0.1267 & 0.0925 & - \\
\midrule
\multirow{5}{*}{KITTI} 
& 8,192  & \bfseries0.1169 & 0.1187 & 0.1630 & 0.1767 \\
& 16,384 & \bfseries0.1114 & 0.1305 & 0.1646 & 0.2095 \\
& 32,768 & \bfseries0.1087 & 0.1663 & 0.1671 & 0.3110 \\
& 65,536 & \bfseries0.1087 & 0.1842  & 0.1674 & - \\
& All    & \bfseries0.1087 & 0.1853  & 0.1674 & -\\
\bottomrule
\end{tabular}
\vspace{-1ex}
\end{table}

\subsection{Generalization results on point density} \label{sec:capacity_and_density_generalization}
We next evaluate how our model generalizes to different point densities.  During training, we sample 8,192 points for each frame. Without any fine-tuning, we evaluate on 16,384, 32,768, 65,536 sampled points. For KITTI, we also evaluate on all points.

Because of our architectural design, we have the advantage of being able to process large-scale point clouds at one time, and thus do not need to divide the scene and feed the parts one by one into the network like \cite{pointnet, pointnet2}. For all our experiments, we feed the two whole point clouds into the network in one pass. The maximum number of points for one frame in KITTI is around 86K.

Table \ref{table:point_density_flyingthings3d} shows the performance of various point densities on both datasets, where we also compare with an identical architecture without our normalization scheme (\textbf{No Norm}). Results show that the normalization scheme has slight information loss. No Norm has best performance on the training density, but our architecture with normalization is the most robust under different densities -- EPE3D does not increase even though we evaluate on totally different point densities from the density used during training.

\subsection{Ablation studies} \label{sec:ablative}
\begin{table}
\centering
\footnotesize
\caption{Ablation studies (EPE3D) on FlyingThings3D. Results show that each component is important.} \label{table:ablative}
\begin{tabular}{@{}cccccc@{ }c@{}}
\toprule
NoSkips & OneCorr & OriNorm & EM & No $Rel.\ Pos.$ & Full\\
\midrule
0.3149 & 0.3698 & 0.6583 & 0.0948 & 0.0989 & \bfseries0.0804\\ 
\bottomrule
\end{tabular}
\vspace{-1ex}
\end{table}

To study the contribution of each component, we conduct a series of ablation studies, where each time we only change one component:
\begin{itemize}
	\item \textbf{NoSkips}: We remove all skip links.	
	\item \textbf{OneCorr}: To validate that using multiple CorrBCLs of different scales improves performance, we only keep the last CorrBCL.
	\item \textbf{OriNorm}: We replace the normalization scheme for each BCL with the original normalization scheme used in previous work~\cite{BCN1, BCN2, splatnet}.	
	\item \textbf{Elementwise Multiplication (EM):} We use elementwise multiplication in patch correlation. Since elementwise multiplication does not support input features of different lengths for the two point clouds, we remove the links from previous CorrBCLs to the next CorrBCLs.
	\item \textbf{No $Rel.\ Pos.$:} We remove all the relative positions that are concatenated with input signals.
\end{itemize}

We see from Table \ref{table:ablative} that the original normalization scheme does not work well for scene flow estimation. Both skip links and multiple CorrBCLs contribute significantly.  
We see that by using concatenation instead of elementwise multiplication, we are able to link previous CorrBCLs to the next CorrBCLs, and thus boost the performance. By taking both global and local positional information, our model obtains improved performance.

\section{Conclusion}

We presented HPLFlowNet, a novel deep network for scene flow estimation on large-scale point clouds.
We proposed the novel DownBCL, UpBCL and CorrBCL and a density normalization scheme, which allow the bulk of our network to robustly perform on permutohedral lattices of different scales. This greatly saves computational cost without sacrificing performance. Through extensive experiments, we demonstrated its advantages over various comparison methods.

\vspace{-10pt}
\paragraph{Acknowledgments.} {This work was supported in part by NSF IIS-1748387, TuSimple and GPUs donated by NVIDIA.} 

{\small
\bibliographystyle{ieee}
\bibliography{egbib}
}

\end{document}